\documentclass{article}

\usepackage[preprint]{neurips_2021}

\usepackage[utf8]{inputenc} 
\usepackage[T1]{fontenc}    
\usepackage{hyperref}       
\usepackage{url}            
\usepackage{svg}
\usepackage{booktabs}       
\usepackage{bm}
\usepackage{wrapfig} 
\usepackage{amsfonts} 
\usepackage{natbib}
\setcitestyle{square,numbers,sort}
\usepackage{amsthm}
\usepackage{amsmath}
\usepackage{multirow}
\usepackage{microtype}      
\usepackage{subfigure}
\usepackage{longtable}
\usepackage{algorithm}
\usepackage{algorithmic}
\usepackage{svg}
\usepackage{ulem} 
\usepackage{bbding}
\usepackage{pgfplots}
\usepackage{pgfplotstable}
\pgfplotsset{width=7.5cm,compat=1.13}

\title{M6-T: Exploring Sparse Expert Models and Beyond}

\author{%
  An Yang$^{*}$, Junyang Lin$^{*}$, Rui Men$^{*}$, Chang Zhou, Le Jiang, Xianyan Jia \\
  \textbf{Ang Wang, Jie Zhang, Jiamang Wang, Yong Li, Di Zhang, Wei Lin, Lin Qu}\\ 
  \textbf{Jingren Zhou, Hongxia Yang$^{\dagger}$}\\
  Alibaba Group\\
  \texttt{\{ya235025,junyang.ljy,menrui.mr,ericzhou.zc,jiangle.jl,} \\ 
  \texttt{xianyan.xianyanjia,wangang.wa,wanglin.zj,jiamang.wang,jiufeng.ly,}\\
  \texttt{di.zhangd,weilin.lw,jingren.zhou,yang.yhx\}@alibaba-inc.com} \\
  \texttt{xide.ql@taobao.com}\\
}

\begin{document}

\maketitle

\begin{abstract}
  Mixture-of-Experts (MoE) models can achieve promising results with outrageous large amount of parameters but constant computation cost, and thus it has become a trend in model scaling. Still it is a mystery how MoE layers bring quality gains by leveraging the parameters with sparse activation. In this work, we investigate several key factors in sparse expert models. We observe that load imbalance may not be a significant problem affecting model quality, contrary to the perspectives of recent studies~\citep{gshard, switch, base}, while the number of sparsely activated experts $k$ and expert capacity $C$ in top-$k$ routing can significantly make a difference in this context. Furthermore, we take a step forward to propose a simple method called expert prototyping that splits experts into different prototypes and applies $k$ top-$1$ routing. This strategy improves the model quality but maintains constant computational costs, and our further exploration on extremely large-scale models reflects that it is more effective in training larger models. We push the model scale to over $1$ trillion parameters and implement it on solely $480$ NVIDIA V100-32GB GPUs, in comparison with the recent SOTAs~\citep{gshard, switch} on $2048$ TPU cores. The proposed giant model achieves substantial speedup in convergence over the same-size baseline. 
  
\end{abstract}

\section{Introduction}
\label{sec:intro}

\renewcommand{\thefootnote}{\fnsymbol{footnote}}
\renewcommand{\thefootnote}{\arabic{footnote}}

Large-scale pretraining has been demonstrating tremendous success across several fields, especially natural language processing~\citep{bert, gpt2, megatron, t5, gpt3}. Recent studies have shown that scaling model size can bring significant quality gains in downstream task performance~\citep{megatron, t5, gpt3}, and the model quality scales as a power law with the data size, model scale, and amount of computation~\citep{scaling_laws}. 
This can be extended to the field of multimodal representation learning~\citep{vilbert, uniter}, where models with outrageous numbers of parameters~\citep{dalle, m6} can achieve outstanding performance in cross-modal understanding and generation. 
However, training dense models is computationally expensive and it is hard to train them on super-large-scale data with limited computational resources. 

Inspired by the success of Mixture-of-Experts (MoE)~\citep{moe, diverse-moe}, some recent studies have focused on training large-scale sparse expert models with high training efficiency~\citep{gshard, switch, m6, fastmoe}. 
An MoE layer consists of multiple experts and thus has a large model capacity. 
Each forward computation routes a token to $k$ experts from $N$, where $k \ll N$. 
Such a routing mechanism allows the combination of data parallelism and expert parallelism. 
Previous studies~\citep{gshard, switch} show that it can achieve obvious performance speedup with the same computational resources. 
However, training such large-scale MoE models can be extremely difficult owing to multiple factors, e.g. system challenges of communication and load imbalance, and algorithmic challenges of training instabilities, etc. 

In this work, we conduct an analysis of the recent MoE models to figure out which factors influence the model quality and training efficiency. 
We investigate several factors concerning sparse expert models, including load balance, top-$k$ routing strategies, MoE attention, etc. 
Our analysis demonstrates that load imbalance is not a significant problem affecting model quality, while the number of sparsely activated experts $k$ and expert capacity $C$ in top-$k$ routing make a great difference in this context, and larger values of $k$ often contribute to better model performance. 
However, with the increase in the value of $k$, training models with conventional top-$k$ routing implementation~\citep{gshard} becomes much less efficient. 

For further exploration, we extend the experiments to large-scale models with over 10 and 100 billion parameters respectively. 
Our findings and proposals are applicable to extremely large-scale models, and results show that expert prototyping has more significant advantages in training larger models.  
To go even further, we push the model scale to over $1$ trillion parameters and successfully implement it on solely $480$ NVIDIA V100-32GB GPUs, compared with the recent SOTAs on $2048$ TPU cores~\citep{gshard, switch}. 
We show that the $1$-trillion-parameter model outperforms the baseline of the identical model scale, and achieves around 5 times of speedup in training convergence. 

To summarize, our contributions are as follows:
\begin{itemize}
    \item We explore key factors inside MoE models, and find that load imbalance may not be a significant problem affecting model quality in this context, while the number of sparsely activated experts $k$ and expert capacity $C$ in top-$k$ routing significantly impact the model performance. 
    \item We propose a simple method called expert prototyping that splits experts into $k$ prototypes and applies $k$ top-$1$ activation with similar efficiency to top-$1$ routing. This strategy improves the model quality but maintains constant computational costs, and it is even more effective in training larger-scale models. 
    \item We advance our models to $1$ trillion parameters and successfully implement it on solely $480$ NVIDIA V100-32GB GPUs, in comparison with the recent SOTAs on 2048 TPU cores. $1$-trillion-parameter expert prototyping model outperforms the same-scale baseline and achieves substantial speedup in convergence. 
\end{itemize}

\section{Sparse Expert Models}
\label{sec:background}
Sparse expert models are regarded as a promising method for model scaling with high training efficiency. 
Training dense models requires extremely high computational costs, while sparse expert models can converge significantly faster as it iterates on more data with far higher efficiency on a time basis. 
Such mode of large-scale model training is also more environmentally friendly. 
In this setting, a large amount of the model weights are distributed to different workers owing to the architecture of MoE layers, and MoE allows the increase in parameters while keeping constant computational costs~\citep{gshard, switch}. 

Mixture-of-Experts is essentially a routing algorithm that routes tokens to $k$ specified experts from $N$ (where $k \ll N$) for forward computation. 
It enables expert parallelism so that experts process input tokens specified by gating functions simultaneously. 
Expert networks, each of which is a multi-layer perceptron, are distributed across workers. We use a gating function
 which specifies $k$ from $N$ experts for an input token representation $x$. The chosen experts process the representation with forward computation and reduce their results with weighted summation based on the gating values:
 \begin{align}
     \tilde{x} = \sum_{i}^{k}p_iE_i(x),\quad p=softmax(g), \quad g&=topk(softmax(W_gx)),
 \end{align}
where $E_i$ refers to the $i$-th expert. The most typical MoE algorithms for large-scale sparse expert models are Switch Transformer and GShard~\citep{switch, gshard}. 
Their key difference lies in their choice of top-$k$ selection, which applies top-$1$ and top-$2$ selection respectively. 
Switch Transformer~\citep{switch} noted that routing a token to $1$ expert only is effective in preserving model quality and reducing computation complexity, contrary to the idea of~\citet{moe} that $k$ should be larger than $1$ so that there are non-trivial gradients to the routing functions. 

What makes a difference in the performance and model quality is the actual implementation of distributed experts. 
Model parallelism allows the partitioning of a large tensor across workers, and our implementation even enables multiple experts on an identical worker, instead of one expert per worker. 
Due to the dynamic nature of top-$k$ routing, it may cause low efficiency if severe load imbalance happens. 
A standard implementation to tackle the problem is the setting of expert capacity~\citep{gshard, switch, meshtf}, which is defined as: 
\begin{align}
    C=\frac{k \cdot T}{N}\cdot\gamma, 
\end{align}
where $T$ refers to the number of tokens in a batch and $\gamma$ refers to the capacity factor which is commonly larger than $1.0$. A larger capacity factor can provide more buffer for expert capacity. 
Tokens are distributed to experts with all-to-all dispatching operations. The token representations processed by selected experts are combined with all-to-all communication to their original workers. 
On the condition of exceeded capacity, token representations skip forward computation by residual connection.
For experts still having space in their capacities after dispatching, we add padding tokens to fill in. Thus the computations and communications costs are positively related to the number of experts $N$ and the expert capacity $C$. 
Low expert capacity can cause a significant amount of dropped tokens if there is load imbalance on experts, but increasing expert capacity will correspondingly enhance the computation and communication costs. 

\section{Exploration of MoE Models}
\label{sec:exploration}

To comprehensively investigate the MoE models, we conduct a series of experiments on multimodal pretraining following the practice of \citet{m6}, and we evaluate different setups, including the routing methods, capacity, MoE attention, etc. Specifically, we pretrain the MoE model on M6-Corpus, and evaluate the model performance on zero-shot image captioning on the dataset E-commerce IC~\citep{m6}. We provide more experimental details in the appendix. 

\subsection{Development of Load Balance}

\begin{figure*}
    \centering
    \includegraphics[width=\linewidth]{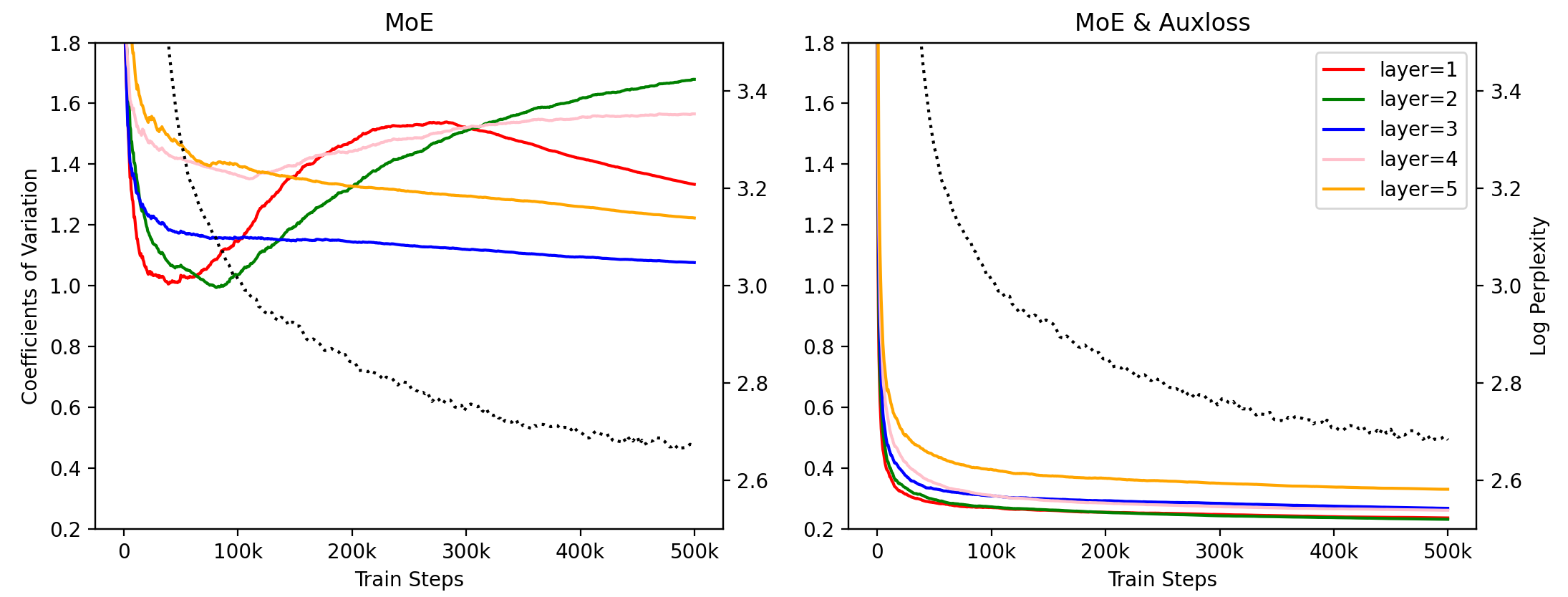}
    \caption{\textbf{Curves of the developments of coefficients of variation $c_v$ at different layers.} Here we demonstrate the development of $c_v$ of baseline and auxiliary loss at all layers. We also demonstrate their curves of training log perplexity (see the black dotted curve). Auxiliary loss helps the model gain highly balanced compute loads at every layer, but the higher balance has not been translated to higher model quality. On the contrary, the behaviors of load balance in the vanilla MoE model are peculiar. Though $c_v$ at all layers drop at the beginning, yet some of them even increase to a high value afterward.}
    \label{fig:load_balance_curve}
\end{figure*}

Recent studies pointed out the significance of balanced routing~\citep{switch, gshard, base}, and illustrated the importance of balancing methods such as auxiliary expert balancing loss. 
We first conduct experiments on the MoE models with and without auxiliary differentiable load balancing loss respectively. 
We pretrain both models for $500k$ steps and compare their upstream and downstream performance. 
Experimental results show that the one with auxiliary loss even performs worse in both upstream evaluation of training log perplexity ($2.694$ vs. $2.645$) and the perplexity evaluation of downstream task of image captioning ($9.97$ vs. $9.72$). 
These observations intrigue us to further investigate the relationship between load balance and model quality. 

We evaluate the degree of compute load balance of experts at every layer. 
Following~\citet{moe}, we also use the coefficient of variation for evaluation. 
We define the coefficient of variation for effective compute loads as that of the number of real tokens computed by the experts $c_v=\frac{\sigma(\mathcal{T})}{\mu(\mathcal{T})}$, 
where $\mathcal{T}$ refers to tokens computed by experts. 
Note that in the implementation, experts that still have buffer should fill padding tokens in their capacities. We do not include this part of computation in the evaluation of compute load balance. This metric reflects the degree of uniformity of token assignment. Furthermore, we focus on the development of $c_v$ in the training process to evaluate the change of load assignment. 

\begin{figure*}
    \centering
    \includegraphics[width=\linewidth]{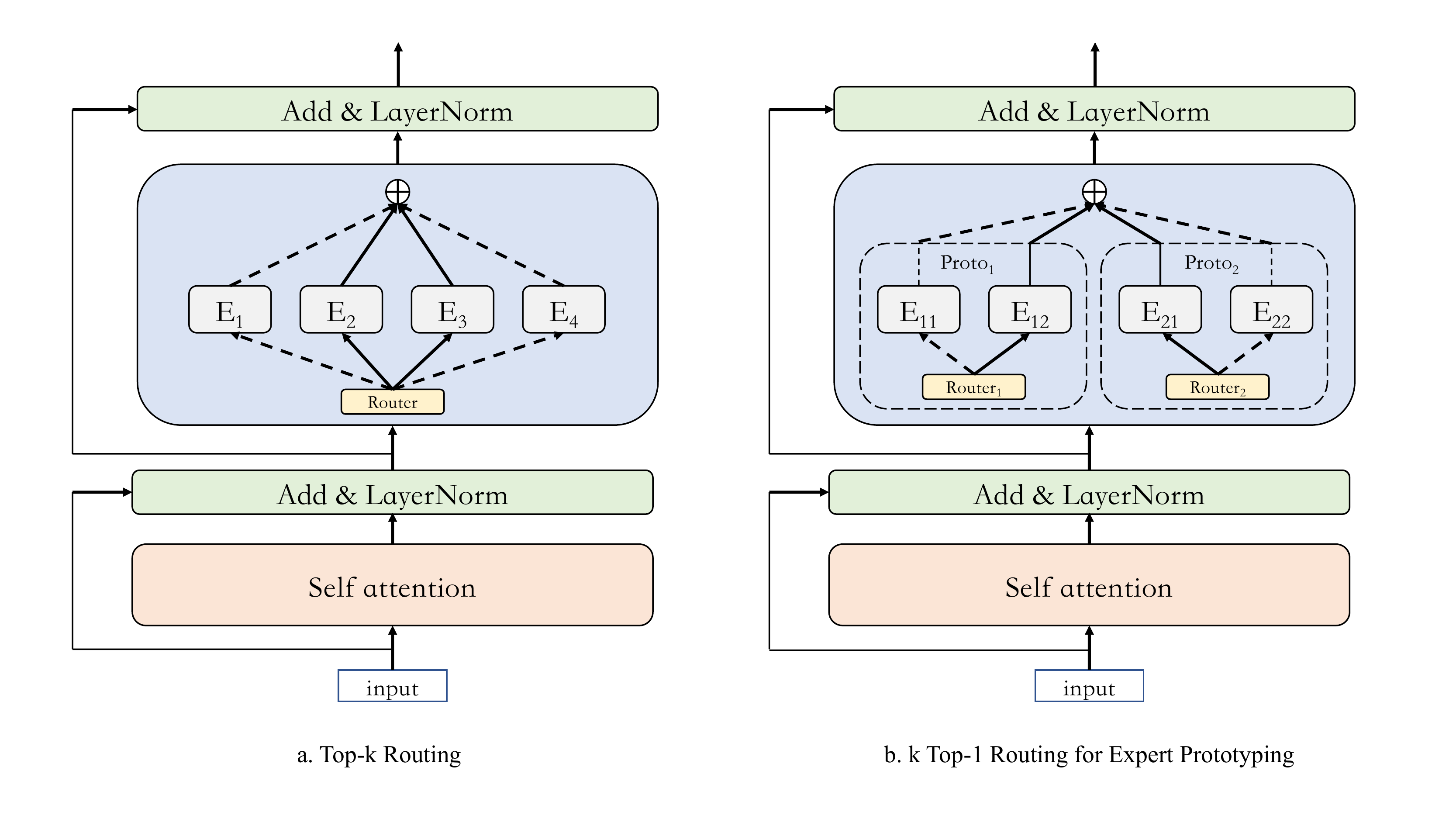}
    \caption{A demonstration of top-$2$ routing and $2$ top-$1$ routing for expert prototyping. In top-$2$ routing, the router selects the top-$2$ from all the experts and sends the token representation to those experts. In $2$ top-$1$ routing for expert prototyping, experts are first grouped into $2$ prototypes, and there is a router for each prototype. Similarly, each router chooses the top-$1$ expert and sends the token representation through it. The output from each prototype is summed up element-wisely for the final output. }
    \label{fig:my_label}
\end{figure*}

We demonstrate the results in Figure~\ref{fig:load_balance_curve}. 
For all layers, significant load imbalance exists at the initial stage according to the high values of $c_v$. 
Notably, the value is generally higher at the top layers. 
For the model trained with auxiliary expert load balancing loss, $c_v$ at all layers drop drastically at the initial stage to a low value around $0.3$ denoting highly balanced compute loads,\footnote{We manually observe the number of tokens that experts receive and find that compute loads are highly balanced} and they become stable in the following. 
However, compute loads are quite different for the MoE model without auxiliary loss. 
Though $c_v$ at all layers drop at the beginning, yet they fail to reach a low one denoting highly balancing compute loads. 
Except for that, some even increase to a high value afterward. 
These phenomena reflect the existence of load imbalance. 
Though auxiliary loss is advantageous in expert load balancing, such an advantage has not been translated to those in upstream and downstream performance, as mentioned above.

\subsection{The Effects of Top-$k$ Sparse Activation}

\begin{figure*}
    \centering
    \includegraphics[width=\linewidth]{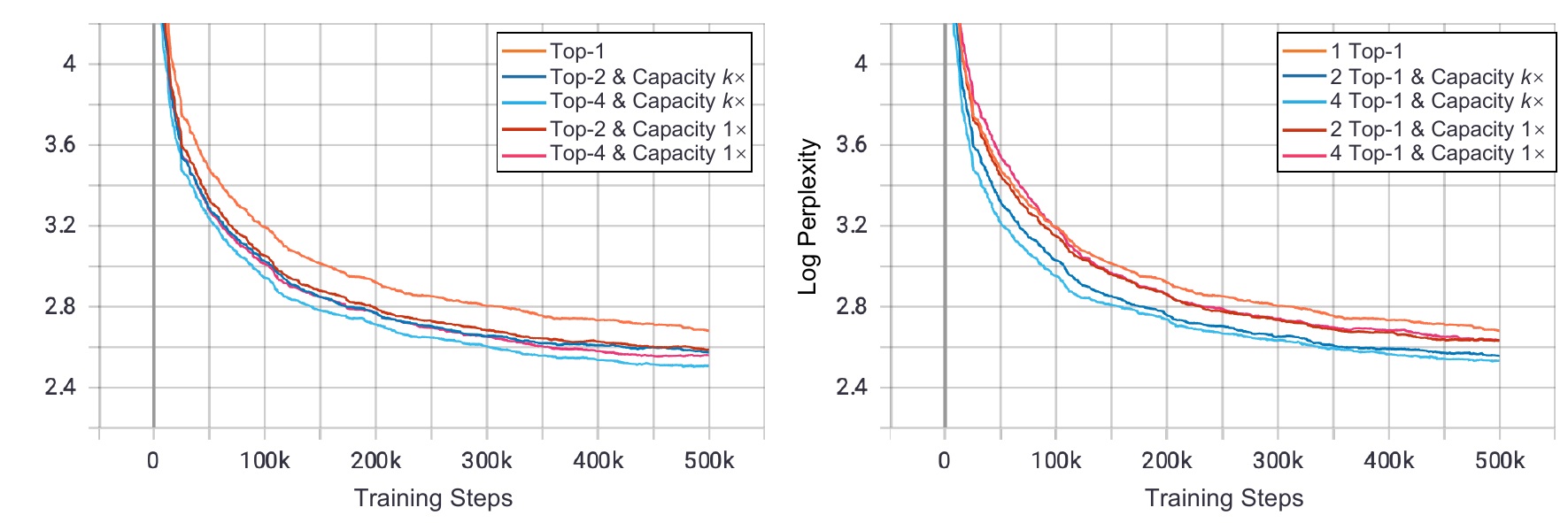}
    \caption{\textbf{Model performance with different top-$k$ routing setups.} The left one demonstrates the performance of top-$k$ routing with different $k$ values. This demonstrates that larger $k$ values even with small capacity bring more benefits, but the gap between top-$2$ and top-$4$ is much smaller than that between top-$1$ and top-$2$. The right one demonstrates the performance of $k$ top-$1$ routing for expert grouping. Similarly, $k$ top-$1$ with $k>1$ routing significantly outperforms the baseline, even with small capacity. }
    \label{fig:topk_ktop1}
\end{figure*}

\begin{table}
\centering
  \caption{\textbf{FLOPs of models with different top-$k$ routing strategies.} We report the computation efficiency (GFLOPs) of top-$k$ routing and $k$ top-$1$ expert grouping with different values of $k$. ``Capacity $k\times$'' refers to the standard setting of expert capacity based on the value $k$, where $C=\frac{kT}{N}\cdot\gamma$, while ``Capacity $1\times$'' refers to the setting of expert capacity of top-$1$ routing for all the models, where $C=\frac{1 \cdot T}{N}\cdot\gamma$. Results show that in the case of limited expert capacity models with different strategies have similar computation FLOPs. } 
  \label{tab:flops}
  \begin{tabular}{ccccccc}
    \toprule
    & Top-$1$ & Top-$2$ & Top-$4$ & $2$ Top-$1$ & $4$ Top-$1$  \\
    \midrule
    Capacity $k\times$    & 2733.53  & 3614.12  & 5619.01  & 3614.00  & 5618.54  \\
    Capacity $1\times$     & 2733.53  & 2733.60  & 2733.67  & 2733.54  & 2733.57  \\
    \bottomrule
  \end{tabular}
\end{table}

Previous work~\citep{switch} pointed out that top-$1$ is sufficient for high model quality in comparison with top-$k$ selection with $k>1$, while it has significant advantages in computational efficiency, but the experimental results in \citet{base} show that top-$2$ still outperforms the top-$1$ selection. Here we conduct experiments to further investigate how top-$k$ methods influence the model quality. 

The value of $k$ determines the number of experts to pass through for each token. We evaluate the model performance on the conditions of $k \in \{1,2,4\}$. 
Note that as mentioned in Section~\ref{sec:background}, the expert capacity $C=\frac{kT}{N}\cdot\gamma$ should be different for different values of $k$, and thus their computation complexities are actually different. 
To evaluate the performance of different routing methods on the condition of identical expert capacity, we experiment on top-$k$ routing methods where $k>1$ with a uniform capacity $C=\frac{1 \cdot T}{N}\cdot\gamma$, denoted by ``Capacity $1\times$'' in the figures and tables. 
In contrast, we denote the original setting of expert capacity $C=\frac{kT}{N}\cdot\gamma$ as ``Capacity $k\times$''. 
We report the computation FLOPs of models with different top-$k$ routing strategies on the conditions of standard and limited expert capacities.\footnote{We report the FLOPs of a single GPU, counted by Tensorflow profiler.} 
Table~\ref{tab:flops} demonstrates that larger values of $k$ lead to higher computation complexity, unless when the expert capacity $C$ is limited. 

The left plot in Figure~\ref{fig:topk_ktop1} demonstrates the effects of choice of $k$ on model convergence. We find that both top-$2$ and top-$4$ sparse activation can outperform the top-$1$. 
Even when $C$ is limited, denoting similar computational complexity, routing methods with $k>1$ still outperform the MoE baseline significantly. 
We assume that tokens can be processed by different experts that function differently. 
However, it is interesting to find the diminishing return that the gap between top-$2$ and top-$4$ is much smaller than that between top-$1$ and top-$2$, especially on the condition of limited expert capacity. 
Besides, due to the looping ``argmax'' operation in top-$k$ routing, the efficiency decreases drastically when $k$ increases, as demonstrated in Table~\ref{tab:speed}. 
It is possible to strike a perfect balance between effectiveness and efficiency by selecting a proper value of $k$. 

\subsection{$k$ Top-$1$ VS. Top-$k$}
\label{subsec:ktop1}
\begin{table}[tb]
\centering
  \caption{\textbf{Speed of models with different top-$k$ routing strategies (ms/step).} We report the training speed of models with different routing strategies. ``Base'' refers to the model with $1.5$ billion parameters, and ``10B'' refers to the large scale model with over $10$ billion parameters. The models are implemented with the limited expert capacity $C=\frac{1 \cdot T}{N}\cdot\gamma$. } 
  \label{tab:speed}
  \begin{tabular}{cccccc}
    \toprule
    & Top-$1$ & Top-$2$ & Top-$4$ & $2$ Top-$1$ & $4$ Top-$1$  \\
    \midrule
    Base     & 214.3  & 218.2  & 305.3  & 220.1  & 225.3   \\
    10B     & 462.2 & 493.0 & 514.2 & 466.9 & 473.9 \\
    \bottomrule
  \end{tabular}
\end{table}
Previous analysis indicates that $k$ experts play different roles and outcompete a single expert. 
However, the looping ``argmax'' operation in top-$k$ routing incurs computation inefficiency. 
Take top-$2$ sparse activation as an example. It performs the operation of ``argmax'' for two times to select the top-$2$ experts. Larger $k$ can incur lower efficiency in the computation. We evaluate the computation efficiency of top-$1$, $2$ and $4$, as shown in Table~\ref{tab:speed}. 
For both ``Base'' model with $1.5$ billion parameters and ``10B'' model with over $10$ billion parameters, top-$k$ routing where $k>1$ especially when $k=4$ is saliently slower than top-$1$.  

To tackle the issue of efficiency, we propose a simple method called expert prototyping, which splits experts into $k$ prototypes. In each forward computation, each token is sent to the $k$ prototypes, and in each prototype, it is processed by the expert in each group selected by top-$k'$ routing where in most cases $k'=1$ for analysis. The outputs of prototypes are combined linearly:
\begin{align}
    y &= \sum_{i=1}^{k}\sum_{j=1}^{m}p_{ij}E_{ij}(x),
\end{align}
where $m$ refers to the number of experts inside a group. 
This method avoids the looping argmax operation, but instead, it generates $k$ outputs in a parallel fashion. 
In comparison with top-$k$, we also name it $k$ top-$1$. 
We find that the increase in $k$ does not cause a significant decrease in training speed, as shown in Table~\ref{tab:speed}. 

We evaluate the model quality of $k$ top-$1$ expert prototyping. 
The right plot of Figure~\ref{fig:topk_ktop1} demonstrates that expert grouping with $k>1$ can achieve better performance compared with the top-$1$ sparse activation. 
However, the improvement in the context of limited expert capacity is much smaller than that in the context of standard expert capacity. 
We find that this problem can be greatly alleviated in training large-scale models, and we leave this issue to Section~\ref{sec:rocket}. 

\subsection{Expert Prototyping and MoE Attention}
\label{subsec:moe_attention}
\begin{figure*}
    \centering
    \includegraphics[width=\linewidth]{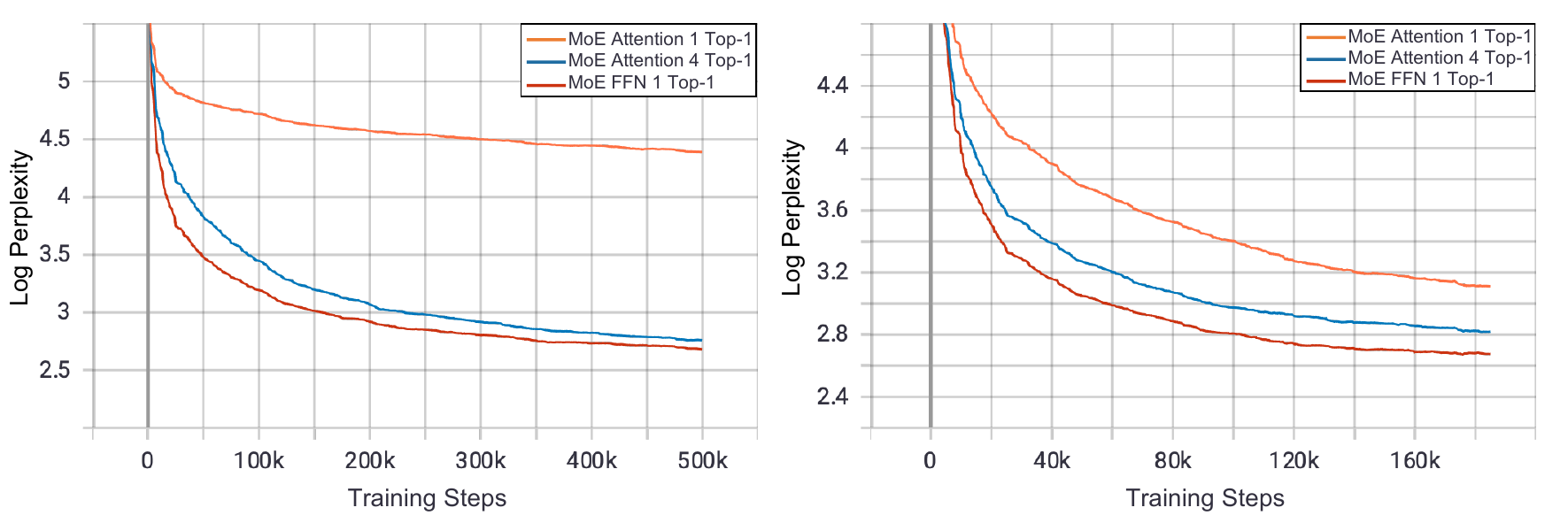}
    \caption{\textbf{Model performance with MoE attention. The left plot and right plot refer to the performance of shallow and deep models respectively.} The left one demonstrates the performance of $k$ top-1 routing with different $k$ values and model design for MoE models. Different from preliminary experiments in \citet{switch}, MoE attention has negative effects on the model performance. The right one demonstrates the performance of MoE attention in the setting of deeper models. Similarly, MoE attention model still performs worse than the MoE baseline. }
    \label{fig:attention_moe_curve}
\end{figure*}

Most work~\citep{moe, diverse-moe,gshard, switch} replaced FFN with MoE.  \citet{switch} pointed out an alternative design of replacing linear transformations in attention with MoE. 
Note that there are $4$ linear transformations for $Q$, $K$, $V$ and output, and they can be viewed as one-layer FFN without non-linear activation. Thus it is possible to replace them with MoE, and we call it MoE attention in this paper.  
Preliminary experiments in~\citet{switch} pointed out that such attention may bring benefits to model quality, but it suffers from training instabilities. 
We implement the MoE attention to our model, and we preserve the MoE layer for FFN. 

Surprisingly, according to the left plot in Figure~\ref{fig:attention_moe_curve}, we reach a different finding that MoE attention negatively affects the model performance, and its training becomes more difficult and even causes divergence. The incorporation of sparse-activated MoE with attention-based correlation computation may highly increase the training difficulty. 
However, $k$ top-$1$ expert prototyping can benefit the training of MoE attention. Expert prototyping with $k>1$ helps the models converge, though the speed of convergence is still slower than that of the MoE baseline. 
Furthermore, we evaluate the effect of MoE attention in the setting of deeper models. Specifically, we increase the number of layers by $4$ times to $20$, but we decrease the number of experts to $8$, so that there is no significant difference in the number of parameters between shallow and deep models. 
We demonstrate the results in the right plot of Figure~\ref{fig:attention_moe_curve}. 
It shows that deeper models with MoE attention perform better and does not diverge, but it still performs worse than the MoE baseline. Still in this case, $k$ top-$1$ expert grouping greatly benefits the model quality, in consistency with the aforementioned observation. 
This shows the effectiveness of $k$ top-$1$ expert prototyping. 

\subsection{Validation on Downstream Tasks}

\begin{table}[]
\centering
\caption{Evaluation of PPL on E-commerce IC \cite{m6}. Following \citet{gpt2}, we do not perform any fine-tuning for any of these results. ``Capacity $k\times$'' refers to the standard setting of expert capacity based on the value $k$, where $C=\frac{kT}{N}\cdot\gamma$, while ``Capacity $1\times$'' refers to the setting of expert capacity of top-$1$ routing for all the models, where $C=\frac{1 \cdot T}{N}\cdot\gamma$.}
\label{tab:downstream_ppl} 

\begin{tabular}{ccccccccc}
    \toprule
 Model Size   & Capacity   & Top-$1$ & Top-$2$ & Top-$4$ & $2$ Top-$1$ & $4$ Top-$1$   \\
    \midrule
Base  &   Capacity $k\times$    & 9.72  & 8.77  & 8.28   &8.73 & 8.39 \\
Base  &   Capacity $1\times$    & 9.72  & 8.84   & 8.52    & 9.24  & 9.41  \\

    \bottomrule

\end{tabular}
\end{table}


We conduct experiments on image captioning and the details of experimental setups are demonstrated in the appendix. 
We demonstrate the experimental results of MoE models with different setups for routing and expert capacity in Table~\ref{tab:downstream_ppl}. 
It can be found that the upstream performance can generally be translated to downstream tasks, which shows that upstream training log perplexity can be an indicator for downstream performance. 
With ``Capacity $k\times$'' expert capacity, both top-$k$ routing and $k$ top-$1$ expert prototyping where $k>1$ significantly outperform the baseline. 
Yet with ``Capacity $1\times$'' limited capacity, $k$ top-$1$ expert prototyping has smaller advantages over the baseline compared with top-$k$ routing. 
This may attribute to the problem of network degeneration. However, this problem is effectively alleviated in training large scale model, as illustrated in Section~\ref{sec:rocket}. 

\section{Rocketing to Trillion Parameters}
\label{sec:rocket}
In this section, we demonstrate that our findings and proposals are also applicable to large-scale pretrained models, and we finally advance the model scale to 1 trillion parameters on solely 480 NVIDIA GPUs, in comparison with the recent SOTA on 2048 TPU cores~\citep{switch}. 

\begin{figure*}
    \centering
    \includegraphics[width=\linewidth]{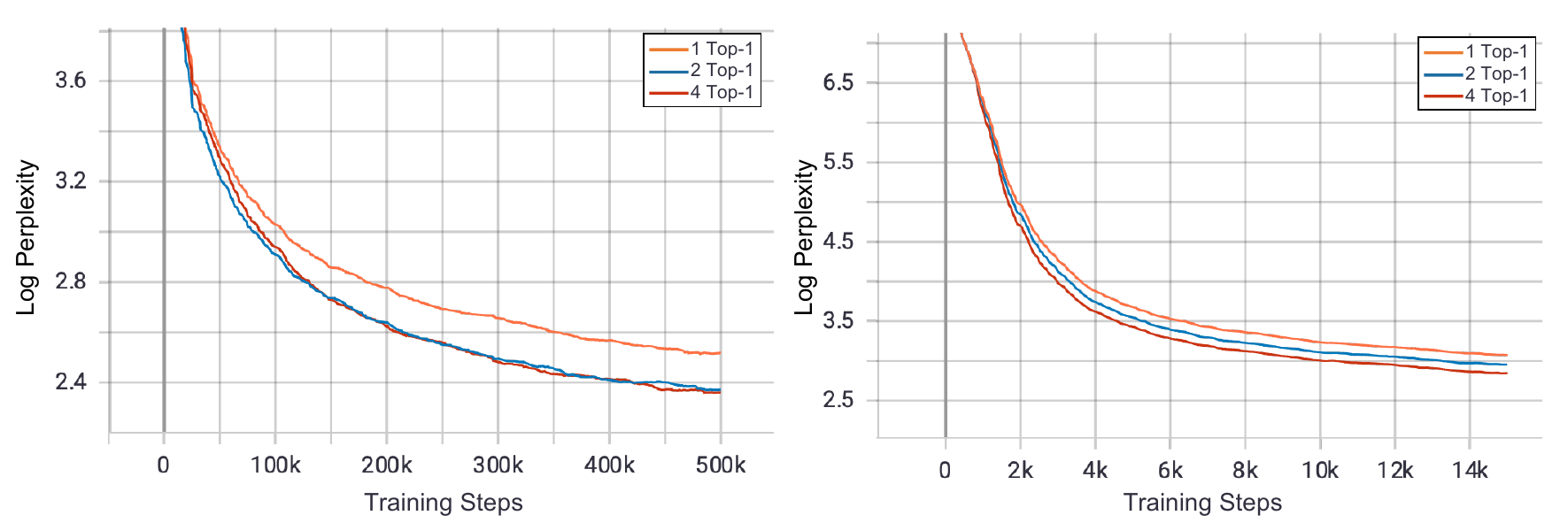}
    \caption{\textbf{Performance of 10B and 100B models with different routing strategies.} In both cases, expert prototyping performs better than the MoE baseline, and larger values of $k$ further benefit model qualities. }
    \label{fig:10b_100b_curve}
\end{figure*}

We validate our findings on extremely large-scale models. We scale the model size up to $10$ and $100$ billion parameters respectively. 
For simplicity, we validate $k$ top-$1$ expert prototyping on the 10B and  100B-parameter models. 
We report the training log perplexity for their upstream performance. 
Figure~\ref{fig:10b_100b_curve} demonstrates their performance, and it can be found that for both 10B and 100B models expert prototyping still have advantages over the MoE baseline, and similarly in both contexts larger $k$ for expert prototyping can further benefit the model quality. 
Besides, as to the downstream performance, Table~\ref{tab:downstream_ppl} shows that for the large scale model ``10B'', $2$ Top-$1$ can reach similar performance to Top-$2$ routing with limited expert capacity. This shows the effectiveness of expert prototyping in training models of a larger scale. 

\begin{table}[]
\centering
\caption{PPL evaluation of ``10B'' models on E-commerce IC \cite{m6}. Note that the ``10B'' models with over $10$ billion parameters are trained with the limited expert capacity where $C=\frac{1 \cdot T}{N} \cdot \gamma$.}
\label{tab:downstream_ppl_10b} 

\begin{tabular}{cccccc}
    \toprule
 Model Size     & Top-$1$ & Top-$2$  & $2$ Top-$1$   \\
    \midrule

10B   & 6.97  & 5.73   & 5.64 \\
    \bottomrule

\end{tabular}
\end{table}


Based on the aforementioned findings and proposals, we move forward to build an extremely large-scale model with over $1$ trillion parameters. 
Due to limited computational resources, we attempt to figure out solutions to implement a $1$-trillion-parameter model on solely $480$ NVIDIA V100-32GB GPUs. 
\begin{figure*}[t]
    \centering
    \includegraphics[width=\linewidth]{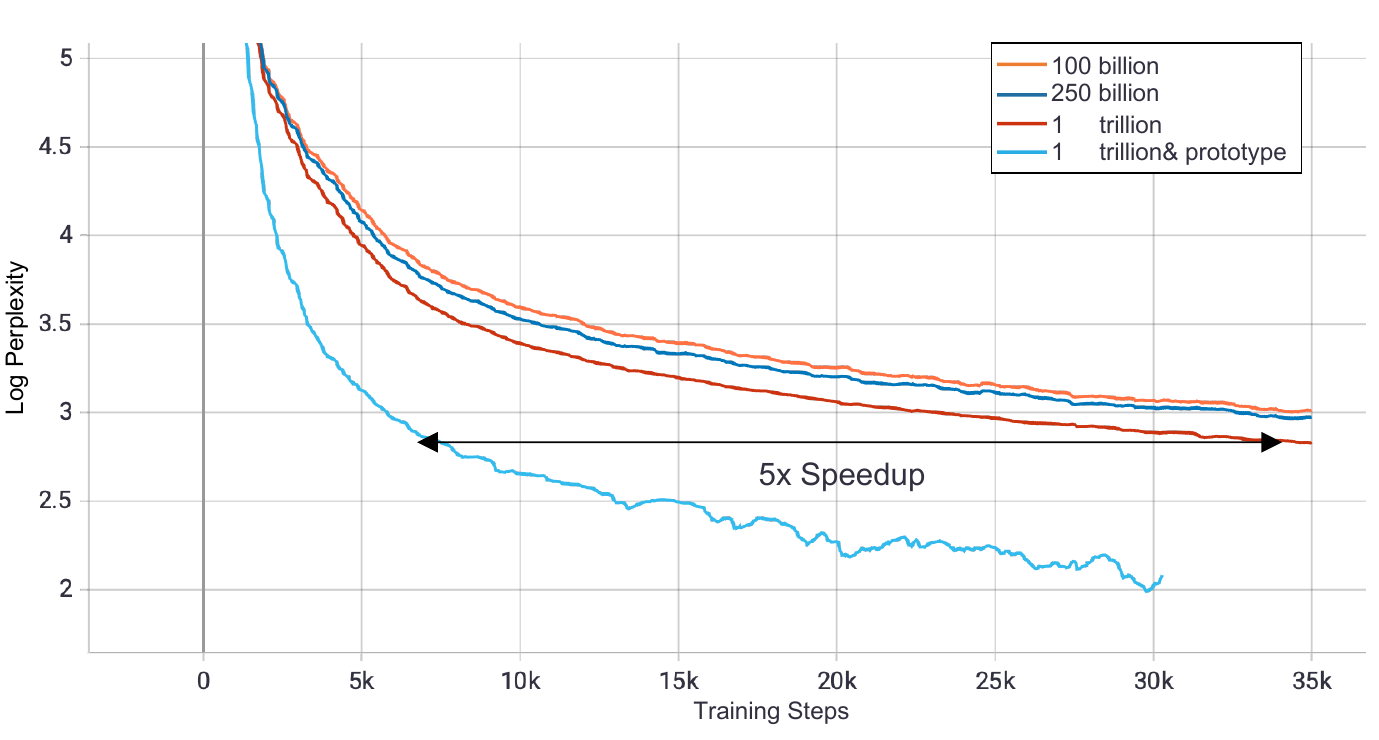}
    \caption{\textbf{Performance of baseline models with $100$ billion, $250$ billion, and $1$ trillion parameters, as well as $1$-trillion-parameter model with expert prototyping.} The curves reflect the scaling law, and also demonstrate the advantage of expert prototyping for giant models.}
    \label{fig:1T_10b_curve}
\end{figure*}

To be more specific, we implement our model on a cluster of single-GPU workers connected by RDMA networks with a bandwidth of 100Gb. 
To save memory usage, we instead turn to Adafactor~\citep{adafactor} for optimization in concern of its sublinear memory costs. 
However, there are a series of sporadic issues concerning training instabilities. 
Through trials and errors, we find that such model training is highly sensitive to learning rates, especially when being trained with Adafactor. 
We did not use the default one $0.01$ due to divergence, but instead, we use $0.005$ to strike a balance between training stability and convergence speed. 
Also, we find that it is essential to lower the absolute values of initialized weights, which is also illustrated in~\citet{switch}. 
We specifically reduce the BERT initialization, a truncated normal distribution with $\mu=0$ and $\sigma=0.02$, by a factor of $10$. 

We first evaluate the quality of models with different parameters but similar computation FLOPs by observing training log perplexity. We compare the performance of MoE baseline models with $100$ billion, $250$ billion parameters, and $1$ trillion parameters, and we observe that the results prove the scaling law that models with larger capacity performs better, as demonstrated in Figure~\ref{fig:1T_10b_curve}.
Then we implement both $1$-trillion-parameter MoE baseline and our expert prototyping MoE model.\footnote{Due to limited computational resources and instabilities in systems and hardware, the trillion-parameter expert prototyping model has been trained for only $30k$ steps. } Still, from Figure~\ref{fig:1T_10b_curve} we can figure out the proposal has a strong advantage over the compared model with $1$ trillion parameters. 
We observe a substantial speedup in convergence, where our method is around $5$ times faster than the baseline. 
However, both models have similar computational FLOPs, which demonstrates that our method strikes a far better balance between computational efficiency and model quality. 

\section{Related work}
\label{sec:related-work}

Pretraining has achieved great success in these years, and it has recently become a common practice in natural language processing~\citep{elmo, bert, gpt, xlnet, roberta, unilm}. In the field of cross-representation learning, pretraining has also become significant and pushed the limit of model performance in downstream tasks~\citep{vilbert, vlbert, vilbert-mt, uniter, villa, oscar, ernie_vil, unimo, vinvl}. 
Recent studies~\citep{scaling_laws} demonstrate the power law of model scale and performance. With the rapid development in distributed training and parallelism~\citep{megatron, zero, zero-offload, zero_infinity}, we have witnessed the burst of studies in extremely large scale pretraining in both natural language processing~\citep{gpt3, megatron} and multimodal pretraining~\citep{dalle, m6} and also new state-of-the-art performance in the recent two years. Though extremely large-scale dense models are highly effective especially in the context of few-shot learning~\citep{gpt3}, some researchers have turned to sparse expert models for efficient large-scale pretraining. Inspired by the success of Mixture-of-Experts~\citep{moe, diverse-moe, meshtf}, recent studies~\citep{gshard, switch} expand the model size to over trillion parameters and fully utilize the advantages of TPUs to build sparse expert models with Mesh-Tensorflow~\citep{meshtf}. They demonstrate that sparse expert models can perform much better than dense models with the same computational FLOPs but their computational costs are similar. A series of the following work successfully implement sparse expert models on NVIDIA GPU~\citep{m6, base}. In this work, we follow the practice of \citet{m6} and implement our models on the distributed learning framework Whale~\citep{whale}.

\section{Conclusion}
\label{sec:conclusion}
In this work, we explore the factors inside sparse expert models and investigate how they influence the model quality and computational efficiency. 
We find out that load imbalance may not be a significant issue affecting model quality, but the number of activated experts $k$ and expert capacity $C$ play a significant role in training MoE models. 
A simple method called expert prototyping that splits experts into different prototypes and applies $k$ top-$1$ routing can help the model achieve improved performance while keeping constant computational costs. 
We extend our series of experiments to extremely large-scale models with over $10$ and $100$ billion parameters and demonstrate the effectiveness of expert prototyping in training extremely large-scale models. 
Finally, we successfully implement the $1$-trillion parameter model on solely $480$ NVIDIA V100-32GB GPUs, compared with the recent SOTAs implemented on $2048$ TPU cores. 
We show that our simple method can improve the performance of $1$-trillion-parameter sparse expert models effectively and help them achieve substantial speedup in convergence.

\bibliographystyle{abbrvnat}
\bibliography{nips}

\clearpage

\appendix

\section{Appendix}
\label{sec:appendix}

\subsection{Multimodal Pretraining and Downstream Evaluation}
In practice, we follow \citet{m6} that employs an extremely large-scale multimodal pretrained model with MoE architecture in Chinese. 
Specifically, we pretrain a model on the image-text pairs from the dataset M6-Corpus~\citep{m6}. In multimodal pretraining, the pretrained model receives the inputs of a pair of related image and text as the input and generates the high-level representations with layers of Transformer~\citep{transformer}. 
In our experiments, we first transform an input image to patch features by splitting it into $4 \times 4$ patches and extracting patch features with a trained ResNet~\citep{resnet}. 
We flatten the patch features of the input image to a sequence of representations and concatenate them with the word embeddings of the text sequence shorter than $128$ words. 
Then we build a feature extractor with multiple layers of transformer consisting of self attention and feed-forward neural networks (FFN). Notably, in order to integrate MoE to the model architecture, we replace the FFN with MoE, where FFN as experts are distributed across workers. 
We pretrain the model with the task of image captioning, where the model learns to generate words autoregressively based on the previous context including the patch features.

To comprehensively evaluate the performance of the methods, we conduct experiments on image captioning in Chinese, and we follow \citet{m6} to use the dataset E-commerce IC. 
We focus on the capability of language modeling of the pretrained model, and thus we use teacher forcing and evaluate the performance by perplexity (PPL).



\subsection{Experimental Setups}
\label{sec:appendix_setups}

For the exploration, we investigate different setups for both models. Here we point out key configurations of our experimental setups and we demonstrate the details in Table~\ref{tab:hparams}. 
Following BERT-Chinese~\citep{bert}, we use the same vocabulary with $21128$ subwords. 
Following the practice of \citet{t5}, we use the same hidden size $1024$ for all the models. We generally scale the model size by increasing the number of layers, the intermediate size, as well as the number of experts. 
For the setup of attention, the number of attention head is $16$ and the attention head size is $64$. 
For the initialization, we use the BERT initialization with $\mu=0$ and $\sigma=0.02$ for most cases, and we use an initialization with a smaller standard deviation of $0.002$ for the $1T$ model. 
As to the expert capacity, we generally use a capacity factor of $\gamma=1.25$ for more buffer. 
The batch size per GPU is $8$ and the total batch size is equal to the product of the batch size per GPU and the number of GPUs. 
We use AdamW optimizer~\citep{adamw} for optimization except for the $1T$ model where we use Adafactor~\citep{adafactor} instead. 
The learning rate for AdamW is $8e-5$, and that for Adafactor is $5e-3$. 
We use warmup schedule with a warmup step of $500$. 
The dropout rate for FFN and attention is $0.1$. 
We use mixed precision training for FP16 communication for all models except the $1T$ one due to the issue of training instability. 

\begin{table}[h]
\centering
  \caption{\textbf{Hyperparameters for pretraining the MoE models.}} 
  \label{tab:hparams}
  \small{
      \begin{tabular}{lcccc}
        \toprule
        \textbf{Hparam} & \textbf{base} & \textbf{10B} & \textbf{100B} & \textbf{1T}  \\
        \midrule
            Hidden size  & 1024 & 1024 & 1024 & 1024 \\
            Intermediate size  & 4096 & 4096 & 4096 & 21248 \\
            Number of layers  & 5 & 10 & 24 & 24 \\
            Number of attention heads  & 16 & 16 & 16 & 16 \\
            Attention head size  & 64 & 64 & 64 & 64 \\
            Initializer range  & 0.02 & 0.02 & 0.02 & 0.002 \\
            Number of experts  & 32 & 128 & 512 & 960 \\
            Number of GPUs  & 8 & 16 & 128 & 480 \\
            Optimizer & AdamW & AdamW & AdamW & Adafactor \\
            Learning rate & 8e-5 & 8e-5 & 8e-5 & 5e-3 \\
            Mixed precision & \checkmark & \checkmark & \checkmark & $\times$ \\ 
            FP16 communication & \checkmark & \checkmark & \checkmark & $\times$ \\ 
            Params & 1.4B & 10.8B & 103.2B & 1002.7B \\ 
        \bottomrule
      \end{tabular}
  }
\end{table}


We implement our experiments on Tensorflow 1.15~\citep{tensorflow}. 
Different from the original implementation of Switch and GShard with Mesh-Tensorflow~\citep{meshtf}, we implement the multimodal pretrained model with the framework Whale~\citep{whale}, which enables data, model, and expert parallelism on NVIDIA GPU.

\subsection{Pseudo Code for Expert Prototyping}
\begin{figure*}
    \centering
    \includegraphics[width=\linewidth]{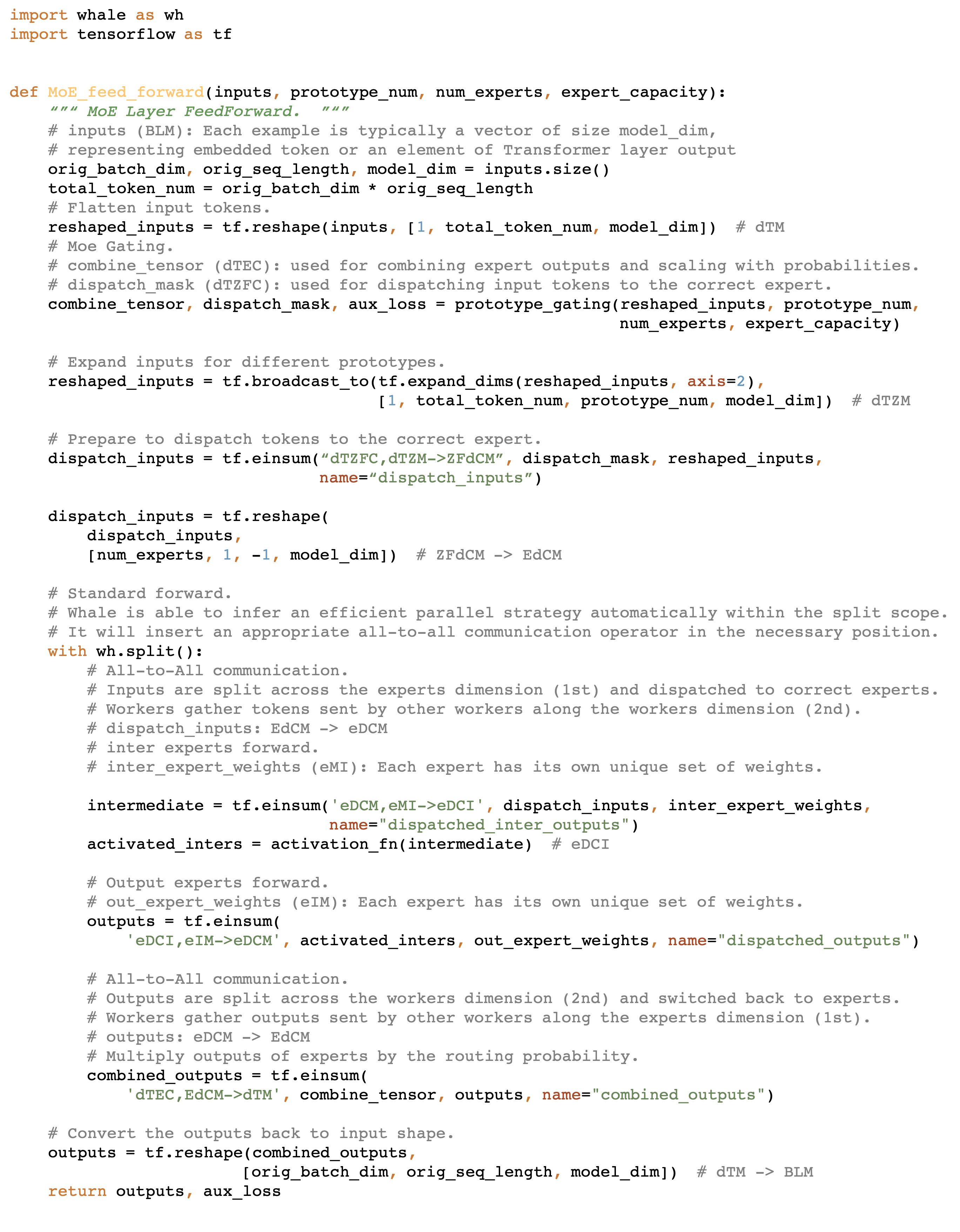}
    \caption{Pseudo code of the MoE Transformer layer in Whale.}
    \label{fig:moe_code}
\end{figure*}

\begin{figure*}
    \centering
    \includegraphics[width=\linewidth]{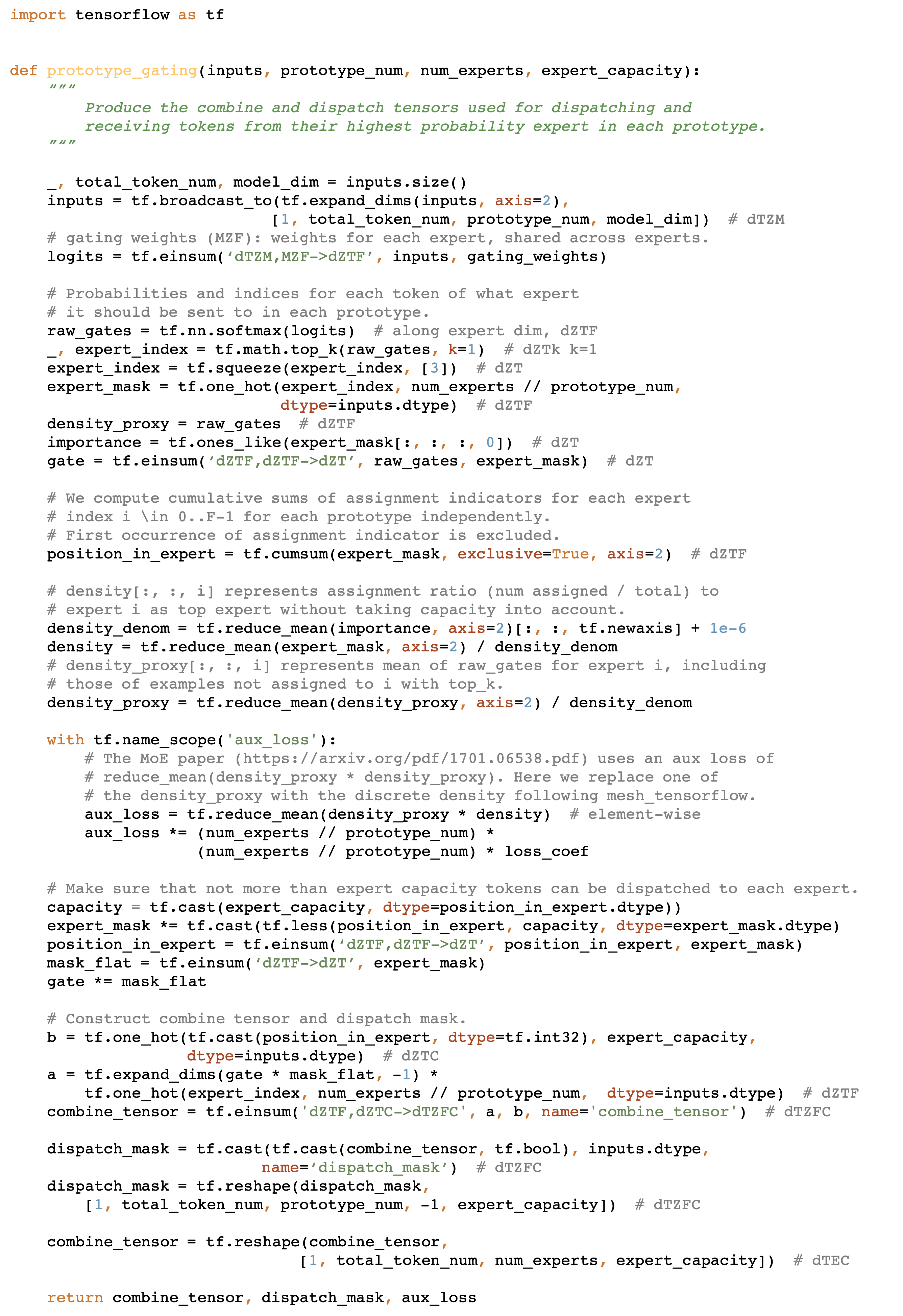}
    \caption{Pseudo code of the Expert Prototyping.}
    \label{fig:prototype_code}
\end{figure*}

The pseudo codes for MoE layer and proposed expert prototyping in Whale are provided in Figure~\ref{fig:moe_code} and Figure~\ref{fig:prototype_code} respectively. 
Table~\ref{tab:notation} illustrates the notations of specific tensor dimensions.
\begin{table}[h]
\centering
\caption{\textbf{Notation table for Pseudo code.}}
\begin{tabular}{cl}
\hline
\textbf{Variable} & \textbf{Definition}                                      \\ \hline
D                 & Number of workers                                        \\
d                 & Number of GPUs per worker (d=1 in this paper)          \\
E                 & Number of total experts                                  \\
e                 & Number of experts per worker (e*D=E)                      \\
C                 & Capacity per expert                                      \\
M                 & Model size (same as hidden size, same as embedding size) \\
I                 & Intermediate size                                        \\
B                 & Batch Size per GPU                                       \\
L                 & Sequence Length                                          \\
T                 & Number of tokens (T=B*L)                                 \\
Z                 & Number of prototypes                                     \\
F                 & Number of expert per prototype (Z*F=E)                   \\ \hline
\end{tabular}

\label{tab:notation}

\end{table}

\paragraph{Amount of All-to-All Communication} There are two operations of all-to-all communication in each MoE FFN layer in a forward propagation process (one for \textit{dispatch\_inputs} and the other for \textit{outputs} in the pseudo code).
During the communication, each entry of the communicated tensor passes to a worker once.
Thus, the total amount of communication, which is $O(EdCM) + O(eDCM) = O(EdCM) = O(ECM)$, depends on the number of experts,  capacity and  model size.

\paragraph{Amount of Computation} The total amount of computation in the MoE FFN layer is mainly dominated by the two matrix multiplications, which transform the input tensor from the hidden size to the intermediate size and then vice versa. The total computation of these two matrix multiplications is $O(DeCMI) + O(DeCIM) = O(ECMI)$. In our profiling on the 1T-scale MoE model, these two operations hold around $98\%$ total forward FLOPs of the MoE FFN layer.

\end{document}